\documentclass{llncs}
\usepackage{amssymb}
\usepackage{amsmath}
\usepackage{graphicx}
\usepackage{hyperref}
\usepackage{url}

\usepackage{float}
\floatstyle{plaintop}
\restylefloat{table}

\newcommand{\keywords}[1]{\par\addvspace\baselineskip
\noindent\keywordname\enspace\ignorespaces#1}

\usepackage{array}
\newcolumntype{C}[1]{>{\centering\let\newline\\\arraybackslash\hspace{0pt}}m{#1}}
\newcolumntype{L}[1]{>{\let\newline\\\arraybackslash\hspace{0pt}}m{#1}}

\begin{document}

\mainmatter  % start of an individual contribution

\title{Overview of ExpertLifeCLEF 2018: how far automated identification systems are from the best experts?}

\titlerunning{LifeCLEF Experts vs. Machine Plant Identification Task 2018}

\author{Herv\'e Go\"eau\inst{1}
  \and Pierre Bonnet\inst{1}
  \and Alexis Joly\inst{2,3}
}

\tocauthor{Herv\'e Go\"eau, Alexis Joly, Pierre Bonnet}

\institute{CIRAD, UMR AMAP, France,
\email{herve.goeau@cirad.fr, pierre.bonnet@cirad.fr}%, http://amap.cirad.fr
\and 
Inria ZENITH team, France, %\\
\email{alexis.joly@inria.fr}%,\\ WWW home page:
%\texttt{http://users/\homedir iekeland/web/welcome.html}
\and
LIRMM, Montpellier, France%\\
%\email{name.surname@inria.fr}%,\\ WWW home page:
%\texttt{http://users/\homedir iekeland/web/welcome.html}
}

\toctitle{LifeCLEF Experts vs. Machine Plant Identification Task 2018}

\maketitle

\begin{abstract}
Automated identification of plants and animals has improved considerably in the last few years, in particular thanks to the recent advances in deep learning. The next big question is how far such automated systems are from the human expertise. Indeed, even the best experts are sometimes confused and/or disagree between each others when validating visual or audio observations of living organism. A picture actually contains only a partial information that is usually not sufficient to determine the right species with certainty. Quantifying this uncertainty and comparing it to the performance of automated systems is of high interest for both computer scientists and expert naturalists. The LifeCLEF 2018 ExpertCLEF challenge presented in this paper was designed to allow this comparison between human experts and automated systems. In total, 19 deep-learning systems implemented by 4 different research teams were evaluated with regard to 9 expert botanists of the French flora. The main outcome of this work is that the performance of state-of-the-art deep learning models is now close to the most advanced human expertise. This paper presents more precisely the resources and assessments of the challenge, summarizes the approaches and systems employed by the participating research groups, and provides an analysis of the main outcomes.
\end{abstract}

\keywords{LifeCLEF, ExpertCLEF, plant, expert, leaves, leaf, flower, fruit, bark, stem, branch, species, retrieval, images, collection, species identification, citizen-science, fine-grained classification, evaluation, benchmark}

\section{Introduction} 
Automated identification of plants and animals has improved considerably in the last few years. In the scope of LifeCLEF 2017 \cite{joly2017lifeclef} in particular, we measured impressive identification performance achieved thanks to recent deep learning models (e.g. up to 90 \% classification accuracy over 10K species). 
This raises the question of how far automated systems are from the human expertise and of whether there is a upper bound that can not be exceeded. A picture actually contains only a partial information about the observed plant and it is often not sufficient to determine the right species with certainty. For instance, a decisive organ such as the flower or the fruit, might not be visible at the time a plant was observed. Or some of the discriminant patterns might be very hard or unlikely to be observed in a picture such as the presence of pills or latex, or the morphology of the root. As a consequence, even the best experts can be confused and/or disagree between each others when attempting to identify a plant from a set of pictures. Similar issues arise for most living organisms including fishes, birds, insects, etc. Quantifying this intrinsic data uncertainty and comparing it to the performance of the best automated systems is of high interest for both computer scientists and expert naturalists. This was the goal of the ExpertCLEF challenge, organized as part of the LifeCLEF 2018 campaign \cite{lifeclef2018}.

%The ExpertCLEF challenge invest this challenge in several points:
 
%\begin{enumerate}
%    \item it extends their result to the plant domain whose specificity is that the available data on the web is scarcer, the risk of confusion higher and, finally, the degree of noise higher.
%    \item it scales the comparison between trusted and noisy training data to 10K of species whereas the trusted training sets used in their study were actually limited to few hundreds of species. 
%    \item it uses a third-party test dataset that is not a subset of either the noisy dataset or the trusted dataset. This allows a more fair comparison. More precisely, the test data is composed of images submitted by the crowd of users of the mobile application Pl@ntNet\cite{joly2016look}. Consequently, it exhibits different properties in terms of species distribution, pictures quality, etc.
%\end{enumerate}

In the following subsections, we synthesize the resources and assessments of the challenge, summarize the approaches and systems employed by the participating research groups, and provide an analysis of the main outcomes.

\section{Dataset}

To evaluate the above mentioned scenario at a large scale and in realistic conditions, we built and shared several different datasets coming from different sources. As training data, we provided all the previous datasets used during the previous PlantCLEF challenge \cite{goeau2017plant}. The test set was built with the best experts in the plant domains, in western Europe. For that test set we created sets of observations that were identified in the field by other experts (in order to have a near-perfect golden standard). These pictures were immersed in a much larger test set that had to be processed by the participating systems.\\
\\
\textbf{Trusted and Noisy data in Training Set \textit{expertclef2018}}: a trusted sub-training set based on the online collaborative Encyclopedia Of Life (EoL). A list of 10K species were selected as the most populated species in EoL data after a curation pipeline (taxonomic alignment, duplicates removal, herbaria sheets removal, etc.). The training set contains 256,287 pictures in total but has a strong class imbalance with a minimum of 1 picture for \textit{Achillea filipendulina} and a maximum of 1245 pictures for \textit{Taraxacum laeticolor}. A noisy sub-training set built through Web crawlers (Google and Bing image search engines) and containing about 1.2 million images. This training set is also imbalanced with a minimum of 4 pictures for \textit{Plectranthus sanguineus} and a maximum of 1732 pictures for \textit{Fagus grandifolia}.\\
\\
The main objective of providing these 2 sub-datasets was to offer to the participants the opportunity to evaluate to what extent machine learning techniques can learn from noisy data compared to trusted data. Pictures of EoL are themselves coming from different sources, including institutional databases as well as public data sources such as Wikimedia, iNaturalist, Flickr or various websites dedicated to botany. This aggregated data is continuously revised and rated by the EoL community so that the quality of the species labels is globally very good. On the other side, the noisy web dataset contains more images but with several types and levels of noise: some images are labeled with the wrong species name (but sometimes with the correct genus or family), some are portraits of a botanist specialist of the targeted species, some are labeled with the correct species name but are drawings or herbarium sheets, etc.\\
\\
\textbf{Pl@ntNet test set}: the test data to be analyzed within the challenge is a large sample of the query images submitted by the users of the mobile application Pl@ntNet (iPhone\footnote{\url{https://itunes.apple.com/fr/app/plantnet/id600547573?mt=8}} \& Androïd\footnote{\url{https://play.google.com/store/apps/details?id=org.plantnet}}). It contains a large number of wild plant species mostly coming from the Western Europe Flora and the North American Flora, but also plant species used all around the world as cultivated or ornamental plants including some endangered species. This test set was obtained after a curation pipeline (collaborative species identification evaluation, author reputation, visual quality evaluation, etc.).
This test set was extended with expert observations, according to the following procedure. First, 125 plants were photographed between May and June 2017, in a botanical garden called the "Parc floral de Paris", and in a natural area located in the north of Montpellier city (southern part of France, close to the Mediterranean sea). The photos have been done with two smartphone models, an iPhone 5 and a Samsung S5 G930F, by a botanist and an amateur under his supervision. The selection of the species has been motivated by several criteria including (i) their membership to a difficult plant group (\textit{i.e.} a group known as being the source of many confusions), (ii) the availability of well developed specimens with well visible organs on the spot and (iii), the diversity of the selected set of species in terms of taxonomy and morphology. About fifteen pictures of each specimen were acquired in order to cover all the informative parts of the plant. However, all pictures were not included in the final test set in order to deliberately hide a part of the information and increase the difficulty of the identification. Therefore, a random selection of only 1 to 5 pictures was operated for each specimen. In the end, a subset of 75 plants illustrated by a total of 216 images related to 33 families and 58 genera was selected.

\section{Task Description}

Based on the previously described testbed, we conducted a system-oriented evaluation involving different research groups who downloaded the data and ran their system. Each participating group was allowed to submit up to 5 \textit{run files} built from different methods (a \textit{run file} is a formatted text file containing the species predictions for all test items). Semi-supervised, interactive or crowdsourced approaches were allowed but had to be clearly signaled within the submission system. But none of the participants employed such methods. The main evaluation metric was the top-1 accuracy.

\section{Participants and methods}
28 participants were registered to the ExpertCLEF challenge 2018. Among this large raw audience, 4 research groups finally succeeded in submitting run files. Details of the used methods and evaluated systems are synthesized below and further developed in the working notes of the participants (CMP \cite{CMP2018expert}, MfN \cite{MfN2018expert}, Sabanci Gebze\cite{Sabanci2018expert}, TUC \cite{TUCMI2018expert}. The following paragraphs give a few more details about the methods and the overall strategy employed by each participant.\\
\\
\textbf{CMP, Dept. of Cybernetics, Czech Technical University in Prague, Czech Republic, 5 runs, \cite{CMP2018expert}}: used an ensemble of a dozen Convolutional Neural Networks (CNNs) based on 2 state-of-the-art architectures (Inception-ResNet-v2 and Inception-v4). The CNNs were initialized with weights pre-trained on ImageNet, then fine-tuned with different hyper-parameters and with the use of data augmentation (random horizontal flip, color distortions and random crops for some models). Each single test image is also augmented with 14 transformations (central/corner crops, horizontal flips, none) to combine and improve the predictions. Still at test time, the predictions are computed using the \textit{Exponential Moving Average} feature of TensorFlow, \textit{i.e.} by averaging the predictions of the set of models trained during the last iterations of the training phase (with an exponential decay). This popular procedure is inspired from Polyak averaging method \cite{polyak1992acceleration} and is known to sometimes produce significantly better results than using the last trained model solely. 
As a last step in their system, assuming that there is a strong unbalanced distribution of the classes between the test and the training sets, the outputs of the CNNs are adjusted according to an estimation of the class prior probabilities in the test set based on an Expectation Maximization algorithm.
The best score of 88.4\% top-1 accuracy during the challenge was obtained by this team with the largest ensemble (CMP Run 3). With half less combined models, the CMP Run 4 reached a close top-1 accuracy and even obtained a slightly better accuracy on the smaller test subset identified by human experts. It can be explained by the strategy during the training of using the trusted and noisy sets: a comparison between CMP Run 1 and 4 clearly illustrates that refining further a model with only the trusted training set after learning it on the whole noisy training set is not relevant. CMP Run 3 which combines all the models seems to have its performances degraded by the inclusion of the models refined on the trusted training set when we compare it with CMP Run 4 on the test subset identified by human experts.\\
\\
\textbf{MfN, Museum fuer Naturkunde Berlin, Leibniz Institute for Evolution and Biodiversity Science, Germany, 4 runs, \cite{MfN2018expert}}: followed quite similar approaches used last year during the PlantCLEF2017 challenge \cite{Mario2017}. This participant used an ensemble of fine-tuned CNNs pretrained on ImageNet, based on 4 architectures (GoogLeNet, ResNet-152, ResNeXT, DualPathNet92), each trained with bagging techniques. Data augmentation was used systematically for each training, in particular random cropping, horizontal flipping, variations of saturation, lightness and rotation. For the three last transformations, the intensity of the transformation is correlated to the diminution of the learning rate during training to let the CNNs see patches progressively closer to the original image at the end of the training. Test images followed similar transformations for combining and boosting the accuracy of the predictions. MfN Run 1 used basically the best and winning approach during PlantCLEF2017 by averaging the prediction of 11 models based on 3 architectures (GoogLeNet, ResNet-152, ResNeXT). However, surprisingly, the runs MfN Run 2 and 3, which are based on only one architecture (respectively ResNet152 and DualPathNet92), performed both better than the Run 1 combining several architectures and models. The combination of all the approaches in MfN Run 4 seems even to be penalized by the winning approach during PlantCLEF2017.\\
\\
\textbf{SabanciU-GTU, Sabanci University, Turkey, 5 runs, \cite{Sabanci2018expert}}: fine-tuned and combined two recent successful CNN architectures: DenseNet (Densely connected convolutional Networks), and SeNet (Squeeze-and-excitation Networks), more precisely a SeNet-ResNet-50. Indeed, SeNet introduces building blocks that can be integrated to any modern CNN such as ResNet-50 and that are designed for improving channel interdependencies by adding parameters to each channel of a convolutional block so that the network can adaptively adjust the weighting of each feature map. For its part, a DenseNet is composed of dense blocks where each unit inside is connected to every unit before it. DenseNet has a counter-intuitive property where fewer parameters than a traditional CNN are required while lessening the vanishing-gradient problem. For the challenge, Sabanci-GTU fine-tuned three pre-trained SeNet-ResNet-50 models and one DenseNet. The two first SeNet-ResNet-50 model were trained only on the trusted dataset, while the third one and the DenseNet were fine-tuned on all the available training datasets. Saliency detection, flip, and several rotation angles were used as data augmentation. SabanciU-GTU run 1, 3, 4 and 5 are various weighted combinations of the outputs of the four fine-tuned models. The best result was obtained by the run 5 by weighting the outputs of the CNNs according to the "quality" and "organ" tags provided in the xml metadata files. Run 3 used also the organ tag with manually fixed weights for giving more weight to pictures with "sexual" organs (flower, fruit) or the entire view. Run 2 applied a Error-Correcting Output Codes approach (ECOC) expressing the 10k classes problem through a n-bits (n = 200 here) error-correcting output code. Each bit is related to a binary classifier splitting arbitrarily and randomly into two sets the 10k species. A binary classifier was a 2-hidden layer shallow networks (500 hidden nodes at each layer) taking as input the features from the last layer of the first SeNet-ResNet-50 trained model. Unfortunately, this approach performed the worst during the challenge.\\
\\
\textbf{TUC MI, Technische Universität Chemnitz, Germany, 5 runs, \cite{TUCMI2018expert}}: this team based their system on three architectures (ResNet-50, Inception-v3 and DenseNet-201) fine-tuned on the noisy or trusted dataset with various data augmentations (horizontal and vertical flip, zooming, rotating, shearing and shifting). DenseNet-201 models was fine-tuned with adjusted class weights over multiple iterations to attempt to balance the classes. The best results was obtained by Run 1 and Run 5 which are ensemble classifiers. Run 1 is based on one ResNet, one Inception-v3 and three DenseNet-201, all fine-tuned with the noisy training dataset, and weighted according to their validation accuracy. Run 5 performed slightly better on the whole test set by using only 3 fine-tuned models instead of 5 in Run 1, (2 ResNet-50 and 1 DenseNet-201) and without a specific weighting rule. 
%\\

\section{Results and "Experts vs. Machines" evaluation}

\begin{figure}%[!t]
\centering
\includegraphics[width=0.95\linewidth]{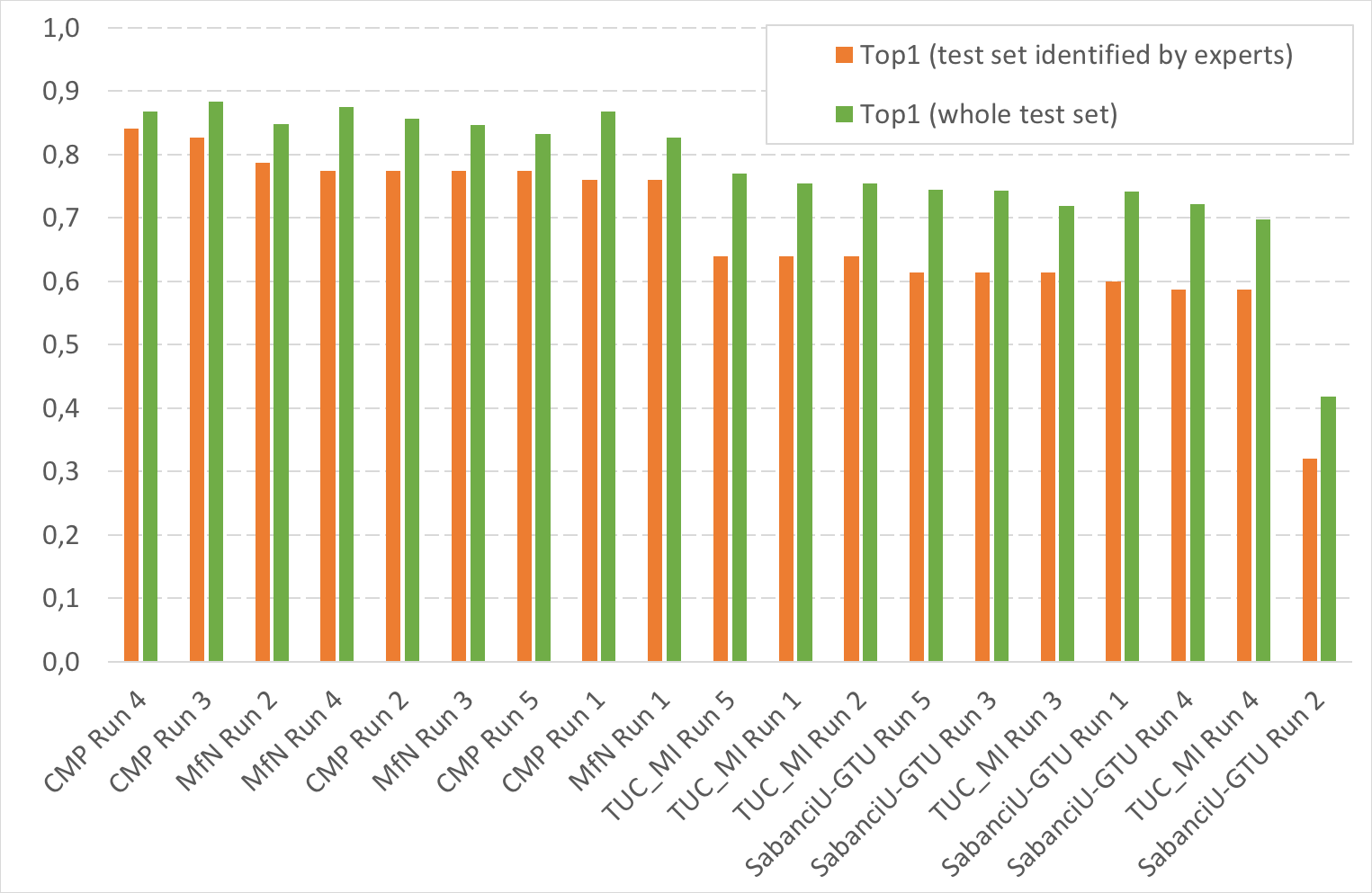}
%\vspace{-10pt}
\caption{Scores achieved by all systems evaluated within the expert identification task of LifeCLEF 2018}
\label{fig:PlantCLEF2018OfficialScore}
\end{figure}

Figure \ref{fig:PlantCLEF2018OfficialScore} and Table \ref{tab:rawresults} provides the performance achieved by the 19 collected runs.\\%. Table \ref{tab:rawresults} provides the results achieved by each run as well as a brief synthesis of the methods used in each of them. 

Considering the automated approaches solely, without comparisons with the experts, we can confirm and remind quickly the same conclusions noticed during the last PlantCLEF 2017 challenge: 
\begin{itemize}
    \item the measured performances are very high despite the difficulty of the task,
    \item the best results were obtained mostly by systems that were learned on both the trusted and the noisy datasets,
    \item all teams used and fine-tuned popular Convolutional Neural Networks confirming definitively the supremacy of this kind of approach over previous methods,
    \item the best results were obtained by ensemble classifiers of ConvNets with many data augmentations.
\end{itemize}

\begin{table}%[h]
    \centering
    \vspace{3mm}
%    \begin{tabular}{|C{25mm}|C{20mm}|C{20mm}|C{20mm}|}
%    \hline
%    Run & Method & Top1 (expert) & Top1 (whole)\\
    \begin{tabular}{|C{35mm}|C{20mm}|C{20mm}|}
    \hline
    Run & Top1 (expert) & Top1 (whole)\\
    \hline
    \hline
CMP Run 4
%& TODO
& 0.840 & 0.867\\
\hline
CMP Run 3
%& TODO
& 0.827 & 0.884\\
\hline
MfN Run 2
%& TODO
& 0.787 & 0.848\\
\hline
MfN Run 4
%& TODO
& 0.773 & 0.875\\
\hline
CMP Run 2
%& TODO
& 0.773 & 0.856\\
\hline
MfN Run 3
%& TODO
& 0.773 & 0.847\\
\hline
CMP Run 5
%& TODO
& 0.773 & 0.832\\
\hline
CMP Run 1
%& TODO
& 0.760 & 0.868\\
\hline
MfN Run 1
%& TODO
& 0.760 & 0.826\\
\hline
TUC MI Run 5
%& TODO
& 0.640 & 0.770\\
\hline
TUC MI Run 1
%& TODO
& 0.640 & 0.755\\
\hline
TUC MI Run 2
%& TODO
& 0.640 & 0.755\\
\hline
SabanciU-GTU Run 5
%& TODO
& 0.613 & 0.744\\
\hline
SabanciU-GTU Run 3
%& TODO
& 0.613 & 0.743\\
\hline
TUC MI Run 3
%& TODO
& 0.613 & 0.718\\
\hline
SabanciU-GTU Run 1
%& TODO
& 0.600 & 0.741\\
\hline
SabanciU-GTU Run 4
%& TODO
& 0.587 & 0.721\\
\hline
TUC MI Run 4
%& TODO
& 0.587 & 0.698\\
\hline
SabanciU-GTU Run 2
%& TODO
& 0.320 & 0.418\\
\hline
\end{tabular}
\caption{Results of the LifeCLEF 2018 Expert Identification Task}%. 
\label{tab:rawresults}
\end{table}

\begin{figure}%[!t]
\centering
\includegraphics[width=0.95\linewidth]{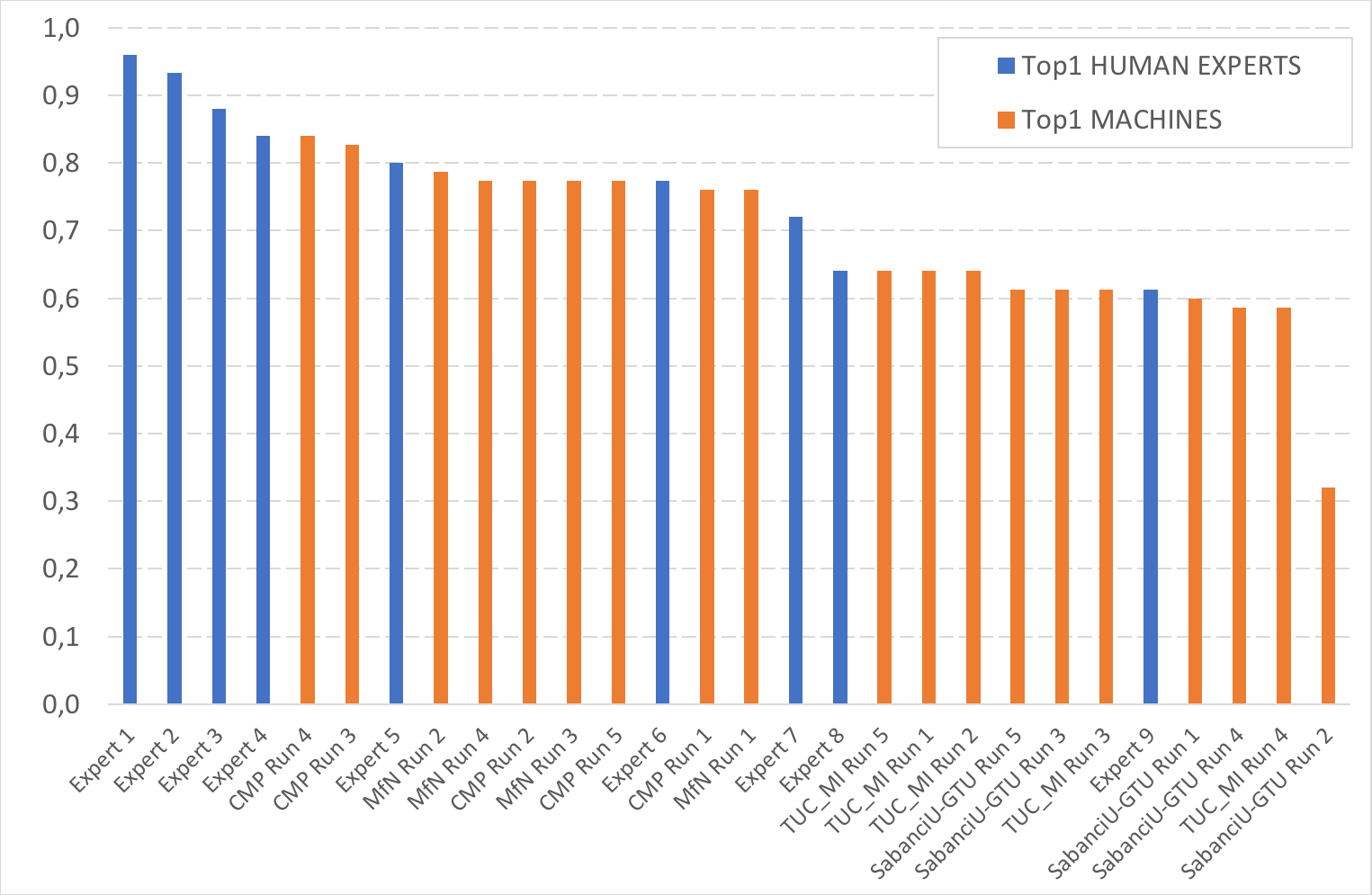}
%\vspace{-10pt}
\caption{Scores between Experts and Machine}
\label{fig:PlantCLEF2018ScoresMvsM}
\end{figure}

Figure \ref{fig:PlantCLEF2018ScoresMvsM} reports the comparison of the Top-1 accuracy between the automated systems and the human experts. The main outcomes we derived from the results of the evaluation are the following ones:\\
\\
\textbf{A difficult task, even for experts:} as a first noticeable outcome, none of the botanist correctly identified all observations. The top-1 accuracy of the experts is in the range $0.613-0.96$. with a median value of $0.8$. This illustrates the difficulty of the task, especially when reminding that the experts were authorized to use any external resource to complete the task, Flora books in particular. It shows that a large part of the observations in the test set do not contain enough information to be identified with confidence when using classical identification keys. Only the four experts with an exceptional field expertise were able to correctly identify more than $80\%$ of the observations.\\
\\
\textbf{Deep learning algorithms were defeated by the best experts} but the margin of progression is becoming tighter and tighter. The top-1 accuracy of the evaluated systems is in the range $0.32-0.84$ with a median value of $0.64$. This is globally lower than the experts but it is noticeable that the best systems were able to perform better than 5 of the highly skilled participating experts. Moreover, regarding a previous Man vs. Machine evaluation in \cite{bonnet2018}, we can notice that some participants succeeded to improve their system in one short year on the same dataset (the best top-1 accuracy was $0.733$ in the previous experiment, $0.84$ during this ExpertCLEF  2018 challenge). We can assume that there is still a room for improvement and that the machines would probably be able to compete with the 3 best human experts next year when the challenge will be re-open on the crowdai platform.\\
\\
\textbf{Identification failures (machines):} looking in details the results, we can notice that some of the best automated systems can perform as well as experts for about 86\% of the observations This is the case for the best evaluated system CMP Run 4 where 65 of the 75 test observations ranked the right species at a lower or equal rank than the best expert. Among the 10 remaining observations, 5 were correctly identified in the top-2 predictions, 2 in the top-3 and only 3 observations were completely failed (see Table \ref{failed-examples-machine}). The causes of the identification failures differs from an observation to another one. For one observation (2792091) it is probably due to a mismatch between the training data and the test sample. Actually, the training samples of the correct species usually contain visible open yellow flowers whereas only beige buds are visible in the test sample. In the second missed observation (2791146), it is more likely that the failure is due to the intrinsic difficulty of the associated genus \textit{Lathyrus} within which many species are visually very similar (but most of the proposals in machine runs are nevertheless under the \textit{Lathyrus} genus). The same for the last missed observation (2791317) related to the genus \textit{Galium} with an additional difficulty related to fact the observation contains only one entire view.\\
%is probably related to the fact that most of images were not focused on a single leaf but dedicated to the illustration of the whole plant, which has a common aspect of a tuft of leaves. The small size of the discriminant organs and the cluttered background in the test sample makes the identification even more difficult. 
\begin{table}%[h]
    \centering
    \begin{tabular}{|c|ccccc|}
        \hline
        Obs. Id & \multicolumn{5}{c}{Photos} \\
        \hline
        2792091 & 
        \includegraphics[width=1.7cm]{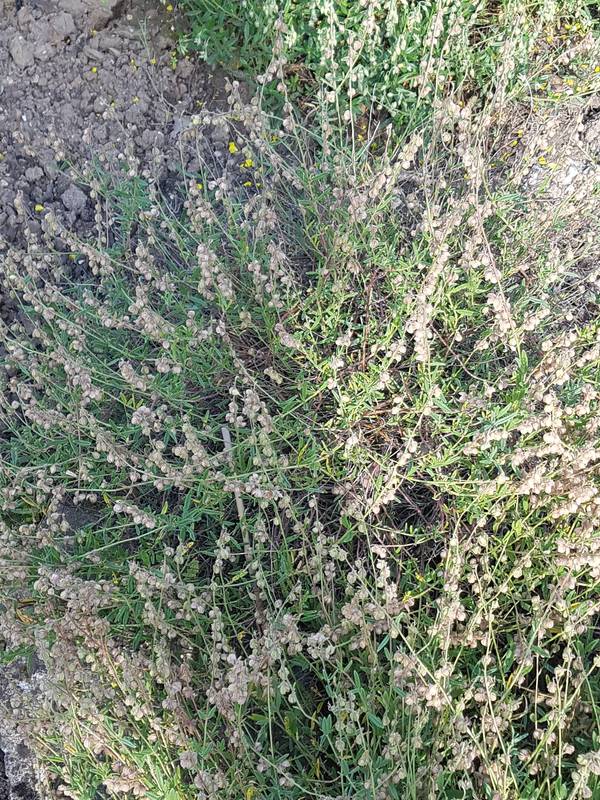} & \includegraphics[width=1.7cm]{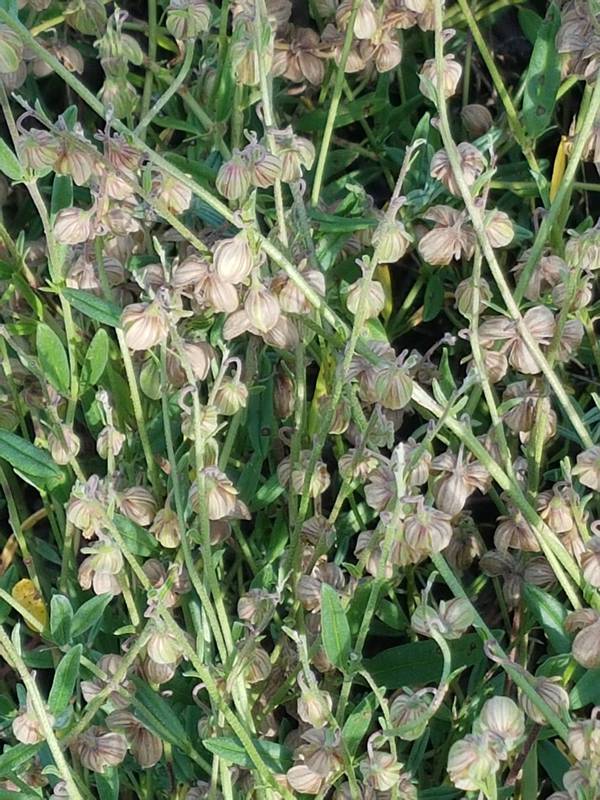} & \includegraphics[width=1.7cm]{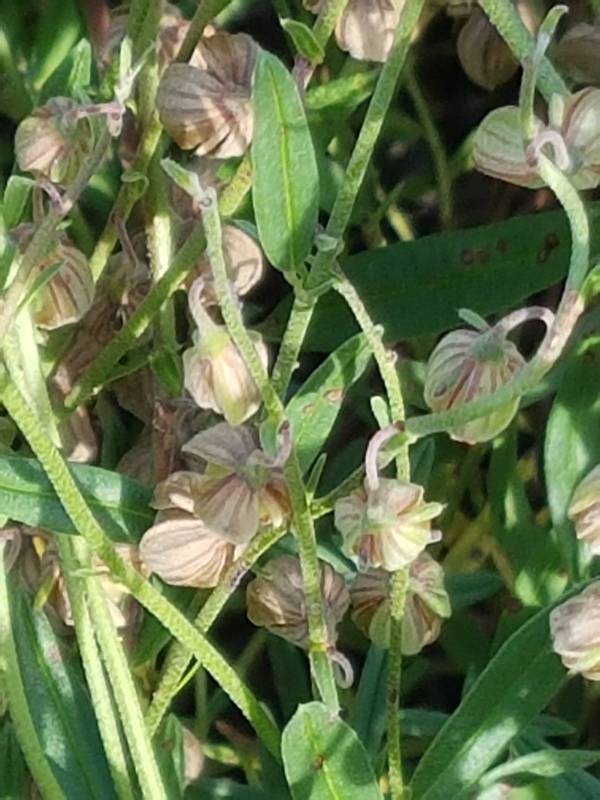} & \includegraphics[width=1.7cm]{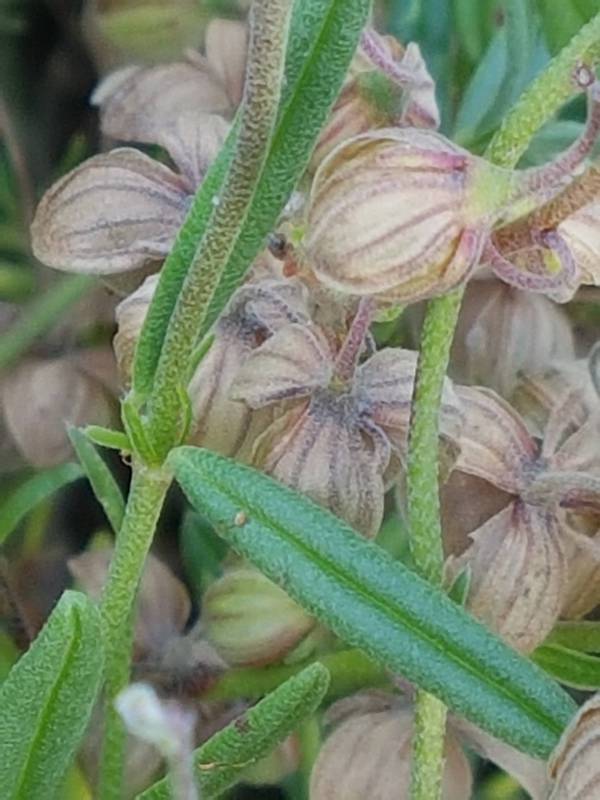} & \includegraphics[width=1.7cm]{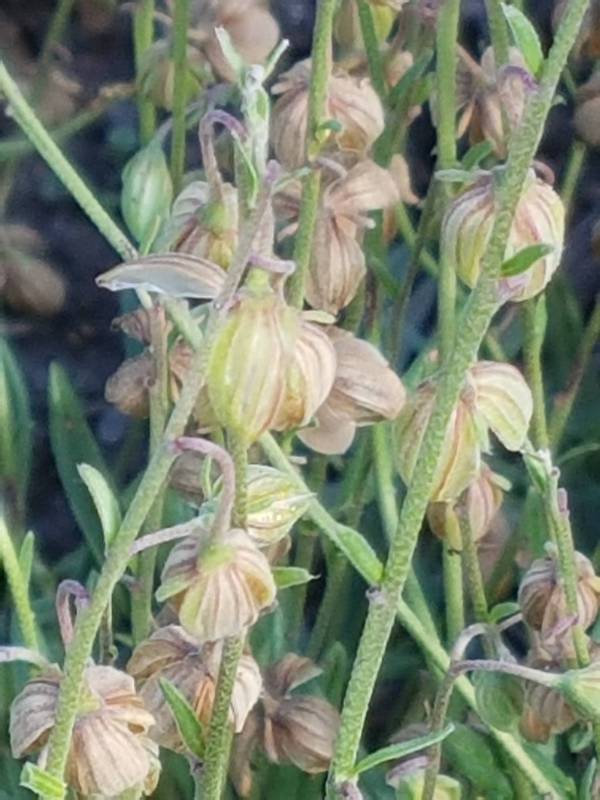} \\
        \hline
        2791146 &
        \includegraphics[width=1.7cm]{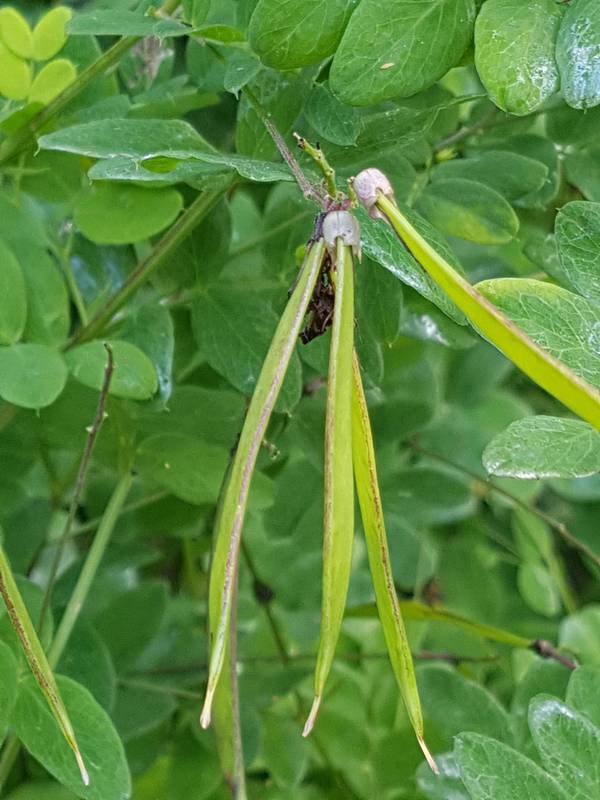} &
        \includegraphics[width=1.7cm]{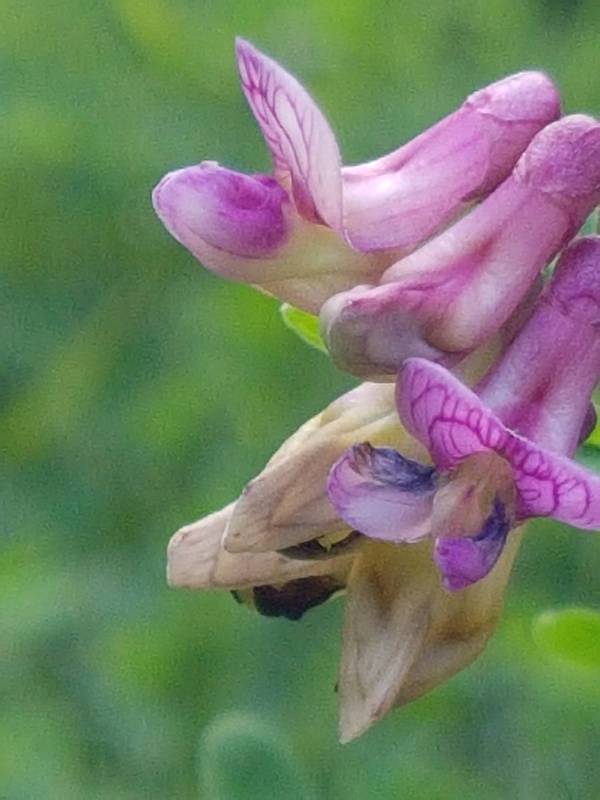} & & & \\
        \hline
        2791317 &
        \includegraphics[width=1.7cm]{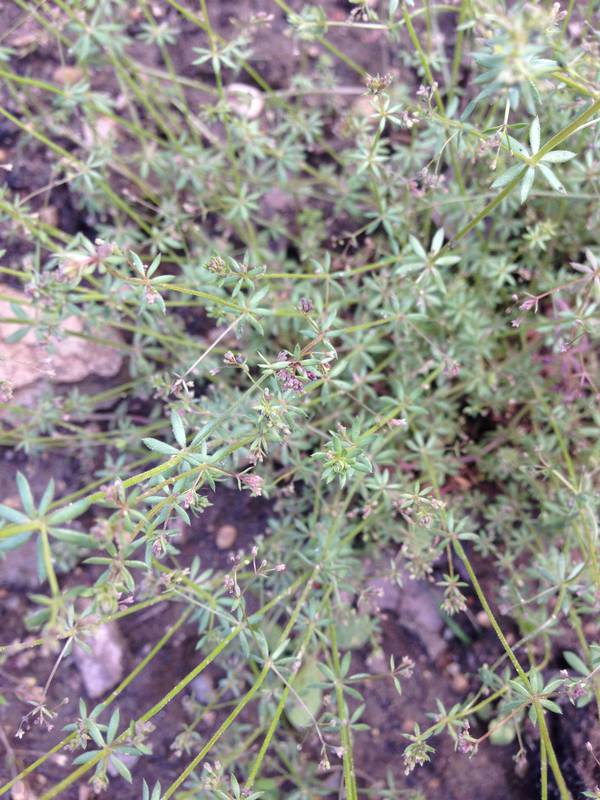} & & & & \\
        \hline
    \end{tabular}
\caption{Examples of observations well identified by experts but missed by the automated identification system.}
\label{failed-examples-machine}
\end{table}
\\
\textbf{Identification failures (experts):} on the other hand, it is important to notice that some automated systems can perform better in some cases than the experts. If we compare again the best automated system CMP Run 4 and the best expert, we can notice that three observations have been better identified by the automated approach (see Table \ref{failed-examples-man}). For one observation (2792706) the best system gave the correct species at rank 1 while it was at rank 2 for the best expert. For the two observations 2790900 and 2791110, the best automated system gave the correct species at rank 3 while there were no species propositions at all from the best human expert. The two observations are actually cultivated plants, probably varieties visually different from the "original" species, and relatively far from the core expertise of the human.
%RELATION AVEC LONGUE TAIL ???

%images illustration 3 observations manquées CMP Run 4 par machine et sur deuxieme ligne 2 observations manquées par Jauzin
%completement  manquées par CMP Run 4 : 2791146, 2791317, 2792091
%completement manquées par Jauzin : 2790900, 2791110, 

%- Jauzin fait mieux que CMP Run 4 pour 10 requetes (5 cas top 1 contre top 2, 2 cas top 1 contre top 3,  3 cas au dela de top 100 pour la machine 

%- CMP Run 4 fait mieux que Jauzin sur 2792706 (bonne espece en position 1 contre 2) et sur 2791110 et  2790900 en top 3 et 5 alors que Jauzin n'a pas repondu. 

%voir si relation avec la longue tail

%One ne peut pas montrer les exemples difficile car revele une partie de la verite terrain

\section{Conclusion}
This paper presented the overview and the results of the LifeCLEF 2018 expert identification challenge following the seven previous LifeCLEF plant identification challenges conducted within CLEF evaluation forum. The task was performed again on the biggest plant images dataset ever published in the literature, but focused on an expert vs. machine evaluation. The main goal behind that was to answer the question of whether automated plant identification systems still have a margin of progression or if they already perform as well as experts for identifying plants in images. We showed that identifying plants from images solely is a difficult task, even for some of the highly skilled specialists who accepted to participate to the experiment. This confirms that pictures of plants only contain partial information and that it is often not sufficient to determine the right species with certainty. Regarding the performance of the automated approaches, we shows that there is still a margin of progression but that it is becoming tighter and tighter. The best system was able to correctly classify $84\%$ of the test samples including some belonging to very difficult taxonomic groups. This performance is still far from the best expert who correctly identified $96.7\%$ of the test samples. However, a strength of the automated systems is that they can return quickly an exhaustive list of all the possible species whereas this is a very difficult task for humans. We believe that this already makes them highly powerful tools for modern botany. Furthermore, the performance of automated systems will continue to improve in the following years thanks to the quick progress of deep learning technologies. They have the potential to become essential tools for teachers and students, but they should not replace an in-depth understanding of botany.

\begin{table}%[h]
    \centering
    \begin{tabular}{|c|ccccc|}
        \hline
        Obs. Id & \multicolumn{5}{c}{Photos}\\
        \hline
        2790900 & 
        \includegraphics[width=1.7cm]{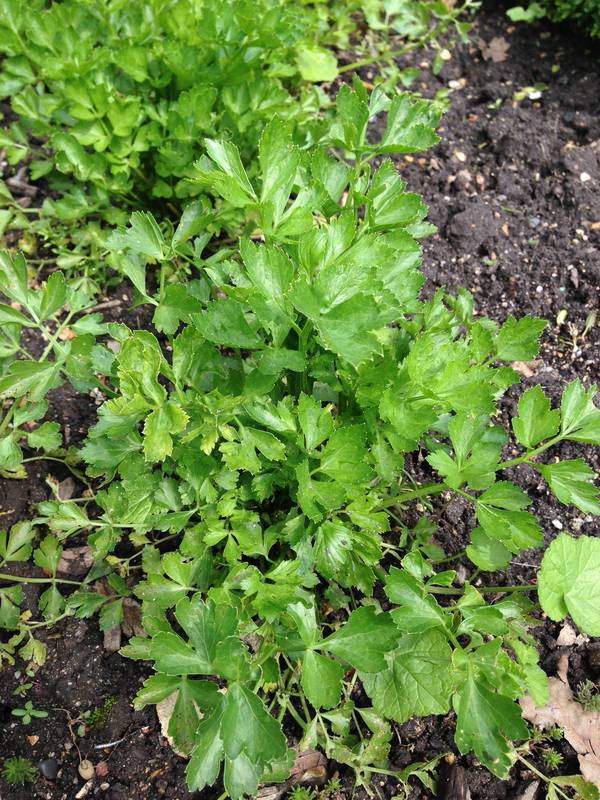} &
        \includegraphics[width=1.7cm]{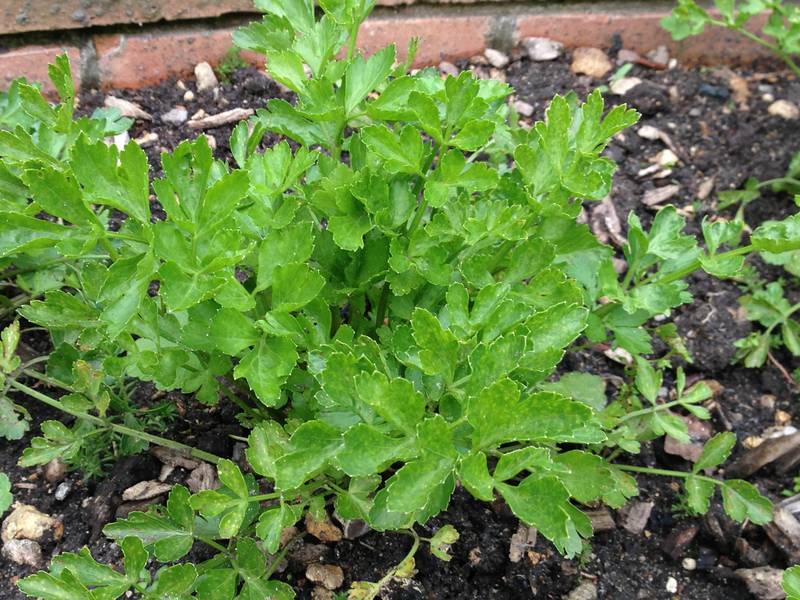} &
        \includegraphics[width=1.7cm]{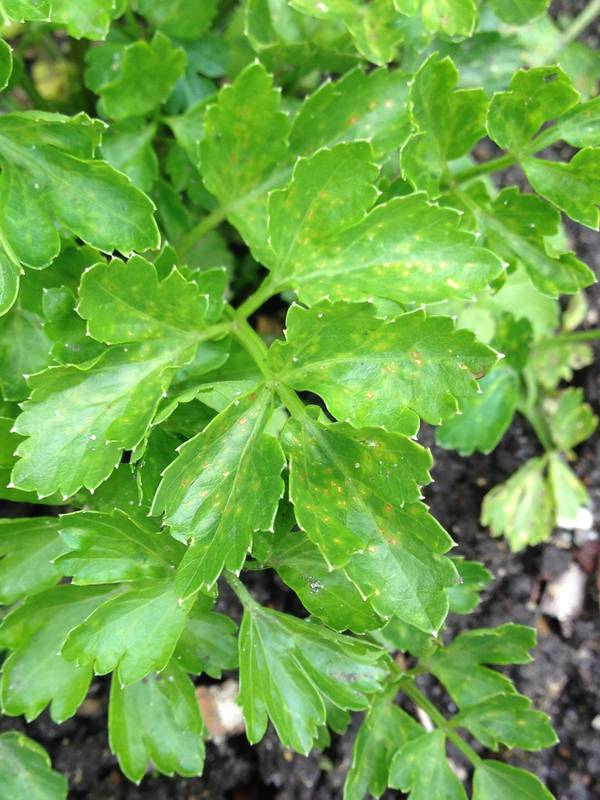} &
        \includegraphics[width=1.7cm]{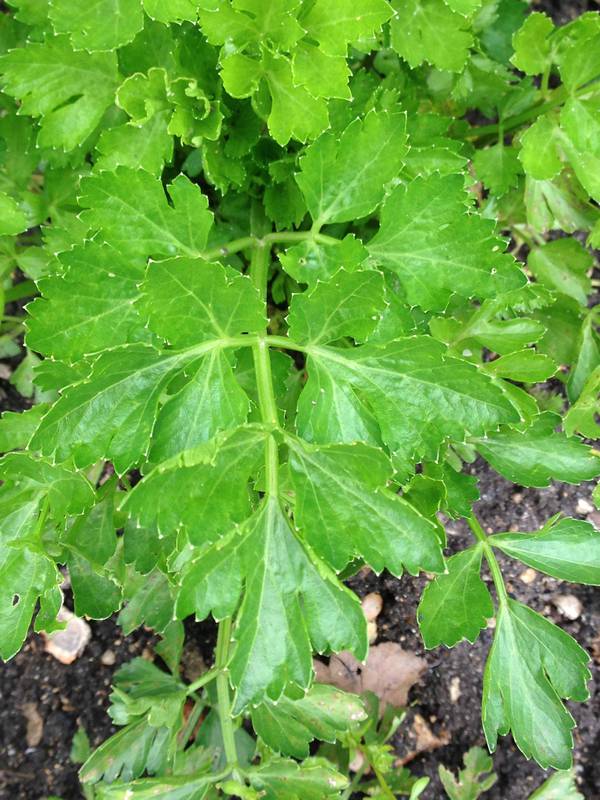} & \\
        \hline
        2791110 &
        \includegraphics[width=1.7cm]{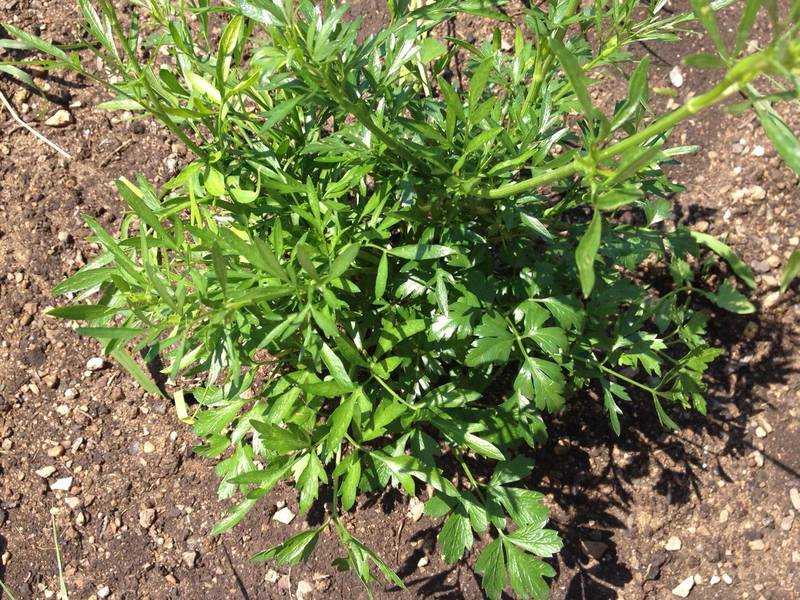} &
        \includegraphics[width=1.7cm]{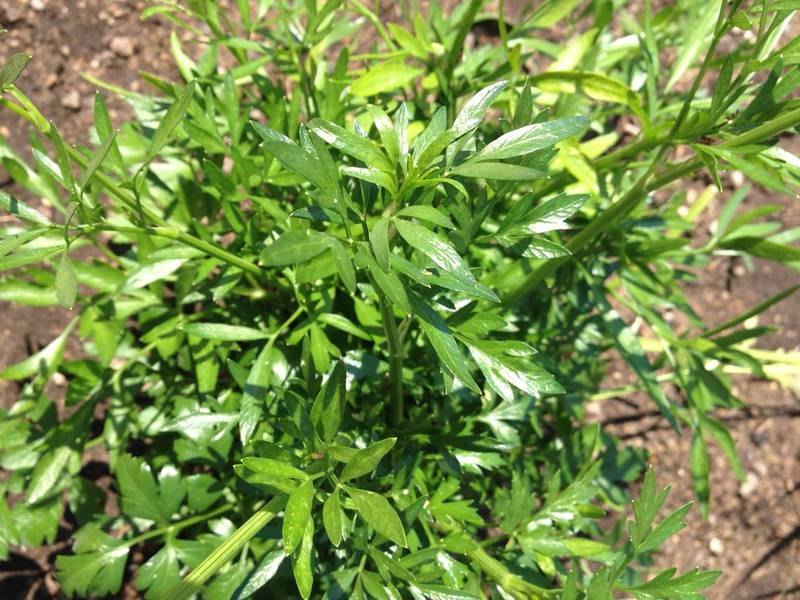} &
        \includegraphics[width=1.7cm]{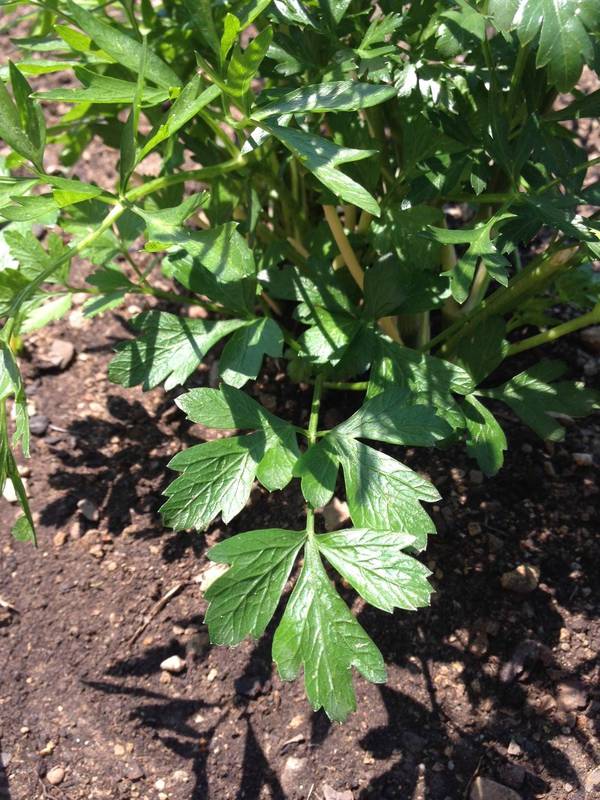} &
        \includegraphics[width=1.7cm]{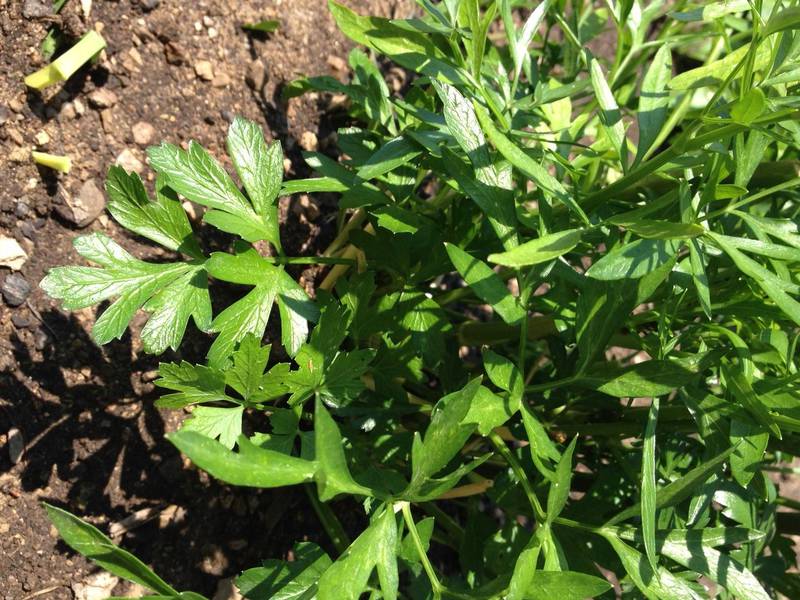} &
        \includegraphics[width=1.7cm]{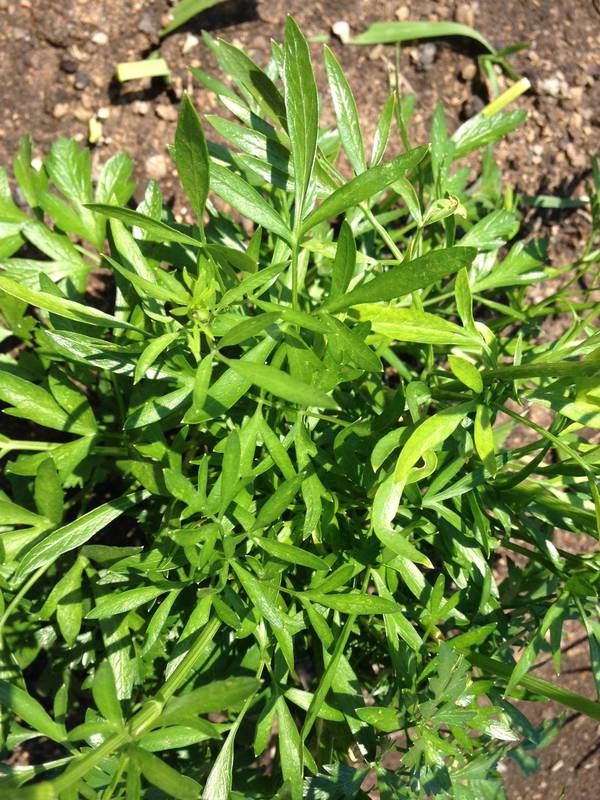} \\
        \hline
    \end{tabular}
\caption{Examples of observations well identified by the automated system but missed by expert.}
\label{failed-examples-man}
\end{table}

\bibliographystyle{splncs03}
%\bibliography{ExpertCLEF2018}

\begin{thebibliography}{10}
\providecommand{\url}[1]{\texttt{#1}}
\providecommand{\urlprefix}{URL }

\bibitem{Sabanci2018expert}
Atito, S., Yanıkoglu, B., Aptoula, E., Ganiyusufoglu, I., Yildiz, A.,
  Yildirir, K., Baris, S.: Plant identification with deep learning ensembles.
  In: Working Notes of CLEF 2018 (Cross Language Evaluation Forum) (2018)

\bibitem{bonnet2018}
Bonnet, P., Go{\"e}au, H., Hang, S.T., Lasseck, M., {\v{S}}ulc, M.,
  Mal{\'e}cot, V., Jauzein, P., Melet, J.C., You, C., Joly, A.: Plant
  Identification: Experts vs. Machines in the Era of Deep Learning, pp.
  131--149. Springer International Publishing, Cham (2018),
  \url{https://doi.org/10.1007/978-3-319-76445-0_8}

\bibitem{goeau2017plant}
Goeau, H., Bonnet, P., Joly, A.: Plant identification based on noisy web data:
  the amazing performance of deep learning (lifeclef 2017). In: CLEF
  2017-Conference and Labs of the Evaluation Forum. pp. 1--13 (2017)

\bibitem{TUCMI2018expert}
Haupt, J., Kahl, S., Kowerko, D., Eibl, M.: Large-scale plant classification
  using deep convolutional neural networks. In: Working Notes of CLEF 2018
  (Cross Language Evaluation Forum) (2018)

\bibitem{lifeclef2018}
Joly, A., Go{\"e}au, H., Botella, C., Glotin, H., Bonnet, P., Vellinga, W.P.,
  M{\"u}ller, H.: Overview of lifeclef 2018: a large-scale evaluation of
  species identification and recommendation algorithms in the era of ai. In:
  Jones, G.J., Lawless, S., Gonzalo, J., Kelly, L., Goeuriot, L., Mandl, T.,
  Cappellato, L., Ferro, N. (eds.) {CLEF: Cross-Language Evaluation Forum for
  European Languages}. Experimental IR Meets Multilinguality, Multimodality,
  and Interaction, vol. LNCS. {Springer}, Avigon, France (Sep 2018)

\bibitem{joly2017lifeclef}
Joly, A., Go{\"e}au, H., Glotin, H., Spampinato, C., Bonnet, P., Vellinga,
  W.P., Lombardo, J.C., Planque, R., Palazzo, S., M{\"u}ller, H.: Lifeclef 2017
  lab overview: multimedia species identification challenges. In: International
  Conference of the Cross-Language Evaluation Forum for European Languages. pp.
  255--274. Springer (2017)

\bibitem{Mario2017}
Lasseck, M.: Image-based plant species identification with deep convolutional
  neural networks. In: Working notes of CLEF 2017 conference (2017)

\bibitem{MfN2018expert}
Lasseck, M.: Machines vs. experts: Working note on the expertlifeclef 2018
  plant identification task. In: Working Notes of CLEF 2018 (Cross Language
  Evaluation Forum) (2018)

\bibitem{polyak1992acceleration}
Polyak, B.T., Juditsky, A.B.: Acceleration of stochastic approximation by
  averaging. SIAM Journal on Control and Optimization  30(4),  838--855 (1992)

\bibitem{CMP2018expert}
Sulc, M., Picek, L., Matas, J.: Plant recognition by inception networks with
  test-time class prior estimation. In: Working Notes of CLEF 2018 (Cross
  Language Evaluation Forum) (2018)

\end{thebibliography}

\end{document}